\pdfoutput=1
\documentclass[11pt]{article}

\usepackage{acl}
\usepackage{times}
\usepackage{latexsym}
\usepackage[T1]{fontenc}
\usepackage[utf8]{inputenc}
\usepackage{microtype}
\usepackage{graphicx}
\usepackage{subfigure}
\usepackage{tipa}
\usepackage{amsmath}
\usepackage{amssymb}
\usepackage{url}
\usepackage{multirow}

\title{Low-Resource Multilingual and Zero-Shot Multispeaker TTS}

\author{Florian Lux \and Julia Koch \and Ngoc Thang Vu \\University of Stuttgart \\\texttt{florian.lux@ims.uni-stuttgart.de}}

\begin{document}
\maketitle 
\begin{abstract}
	While neural methods for text-to-speech (TTS) have shown great advances in modeling multiple speakers, even in zero-shot settings, the amount of data needed for those approaches is generally not feasible for the vast majority of the world's over 6,000 spoken languages. In this work, we bring together the tasks of zero-shot voice cloning and multilingual low-resource TTS. Using the language agnostic meta learning (LAML) procedure and modifications to a TTS encoder, we show that it is possible for a system to learn speaking a new language using just 5 minutes of training data while retaining the ability to infer the voice of even unseen speakers in the newly learned language. We show the success of our proposed approach in terms of intelligibility, naturalness and similarity to target speaker using objective metrics as well as human studies and provide our code and trained models open source.
\end{abstract}

\section{Introduction}
The applications of modern TTS systems are omnipresent and bring major benefits in a very diverse range of tasks. For example, low-resource TTS can be used to revitalize and conserve languages with diminishing numbers of speakers \cite{pine2022requirements}. Other recent applications go into the direction of protecting the privacy of a speaker, by exchanging their voice for a different voice, while not affecting the content of what is said \cite{anonymization}. Even in literary studies, TTS systems can be applied to investigate perceptive aspects of poetry reading \cite{PoeticTTS}. However, while the first of those examples can be done with just a single speaker, the latter two require the TTS system to be able to exchange the voice of the utterance that is produced, which usually requires large amounts of clean multispeaker data. The same requirement exists for many other such applications, which can also be seen in the rise of interest in the research community on voice-cloning technologies \cite{Wu2022AdaSpeech4A, casanova2022yourtts, Neekhara2021AdaptingTM, Hemati2021ContinualSA, cooper2020zero}. 
The communities of speakers of low-resourced languages are thus mostly locked out of plenty of the applications that modern TTS enables. For many instances of such languages, like the Taa language, which is famous for its 83 click sounds or the Yoruba language, in which the tones bear so much meaning, that the language can be mostly whistled, it would be extremely difficult to collect the required amounts of data, and transfer learning to such unique languages is very challenging. Still, we believe that a single model that speaks many languages with any voice can exhibit strong generalizing properties and is a promising first step towards fixing these inequalities.

In this work we ask the following question: Can a multilingual TTS system be used to achieve zero-shot multispeaker TTS in a low-resource scenario? Our approach is to use crosslingual knowledge-sharing to enable 1) finetuning a TTS on just 5 minutes of data in an unseen language in an unseen branch in the phylogenetic tree of languages and 2) transferring zero-shot multispeaker capabilities from the pretraining languages to the unseen language. To achieve this, we propose changes to a TTS encoder to better handle multilingual data and disentangle languages from speakers. Further, we show that the LAML pretraining procedure \cite{finn2017modelagnostic, lux2022laml} can also be used to train general speaker-conditioned models. To verify the effectiveness of our contributions, we train models on just 5 minutes of German and Russian while excluding all Germanic and Slavic languages from the pretraining respectively. We choose a simulated low-resource scenario over an actual low-resource scenario in order to get more reliable evaluations using both objective measures as well as human studies. Furthermore, we show that models trained with this approach do not only serve as a basis for low-resource finetuning with greatly reduced data-need, they can also be used without finetuning as strong multispeaker and multilingual models. We train a model on 12 languages simultaneously and show that it can transfer speaker identities across all languages, even the ones where it has only seen a single speaker during training.

 All of our code, as well as the trained multilingual model are available open source\footnote{\url{https://github.com/DigitalPhonetics/IMS-Toucan}}. An interactive demo\footnote{\url{https://huggingface.co/spaces/Flux9665/IMS-Toucan}} and a demo with pre-generated audios\footnote{\url{https://multilingualtoucan.github.io/}} are available.

\section{Related Work}
\subsection{Zero-Shot Multispeaker TTS}
Zero-shot multispeaker TTS has first been attempted in \cite{arik2018neural}. The idea of using an external speaker encoder as conditioning signal was further explored by \cite{jia2018transfer}. \cite{cooper2020zero} attempted to close the quality gap between seen and unseen speakers in zero-shot multispeaker TTS using more informative embeddings. With the use of attentive speaker embeddings for more general speaking style encoding \cite{wang2018style, choi2020attentron} as well as different decoding approaches in the acoustic space such as generative flows \cite{casanova2021sc}, further attempts have been made at closing the quality gap between seen and unseen speakers. This is however still not a fully solved task. Furthermore, zero-shot multispeaker TTS requires a large amount of high quality data featuring many different speakers to cover a variety of voice properties. 

\subsection{Low-Resource TTS}
In some languages, even a single speaker TTS is not feasible due to the severe lack of high-quality training data available. Attempts at enabling TTS on seen speakers in low-resource scenarios have been made by \cite{9208651, 3403331, tu2019endtoend} through the use of transfer learning from multilingual data, which comes with a set of problems due to the mismatch in the input space (i.e. different sets of phonemes) when using multiple languages. Training a model jointly on multiple languages to share knowledge across languages has been attempted by \cite{he2021multilingual, dekorte2020efficient, yang2020towards}. One solution to the problem of sharing knowledge across different phonemesets is the use of articulatory features, which has been proposed in \cite{Staib_2020, dan, lux2022laml}.

\subsection{Multilingual Multispeaker TTS}
The task of multilingual (not even considering low-resource languages) zero-shot multispeaker TTS is mostly unexplored. YourTTS \cite{casanova2022yourtts} claims to be the first work on zero-shot speaker transfer across multiple languages and was developed concurrently to this work. At the time of writing, there is only a preprint available, so our comparison to their model and methods may differ to a later version. YourTTS reports similar results to ours on high-resource languages using the VITS architecture \cite{kim2021conditional} with a set of modifications to handle multilingual data. The authors find that their model doesn't perform as well with unseen voices in languages that have only seen single speaker training data. Through the low-resource focused design, our approach does not exhibit this problem, while being conceptually simpler. It is shown that just one minute of data suffices to achieve very good results in adapting to a new speaker in a known language with YourTTS. This is consistent with our results, however we go one step further and show that 5 minutes of data is enough to not only adapt to a new speaker, but also to a new language. Also consistent with their results we see that the speaker embedding learns to attribute noisy training data to certain speakers, so not all speakers perform equally well. Ideally we would want to also disentangle the noise modeling from the speakers and languages. The GST approach \cite{wang2018style} has shown that disentangling noise from speakers is possible, it is however not trivial to also disentangle languages, since language properties are also relevant to the encoder, not only the decoder. 

Finally, combining the task of zero-shot multispeaker TTS with the task of low-resource TTS has to the best of our knowledge only been attempted once in a very recent approach that was developed concurrently to ours \cite{9673749}. Their system uses a multi-stage transfer learning process, that starts from a single speaker system which is expanded with a pretrained speaker encoder. They add the required components for speaker and language conditioning and apply finetuning to only those parts of the architecture. The main difference of our system to theirs is that we train the full architecture jointly on the high-resource source domain using the LAML pretraining procedure.

\section{Proposed Method}

\subsection{System Architecture}

Due to its elegant solution to the one-to-many problem of speech synthesis, we choose FastSpeech 2 \cite{ren2020fastspeech} as the basis for our method. 
There is however no reason why this procedure should not work in conjunction with any comparable architecture, making the approach mostly model agnostic. 

We use the Conformer architecture \cite{gulati2020conformer} in both encoder and decoder. This is the same as the basic implementation in the IMS Toucan toolkit \cite{lux2021toucan} which is in turn based on the ESPnet toolkit \cite{hayashi2020espnet, hayashi2021espnet2}.  

\begin{figure}[t]
	\centering
	\includegraphics[width=.7\linewidth]{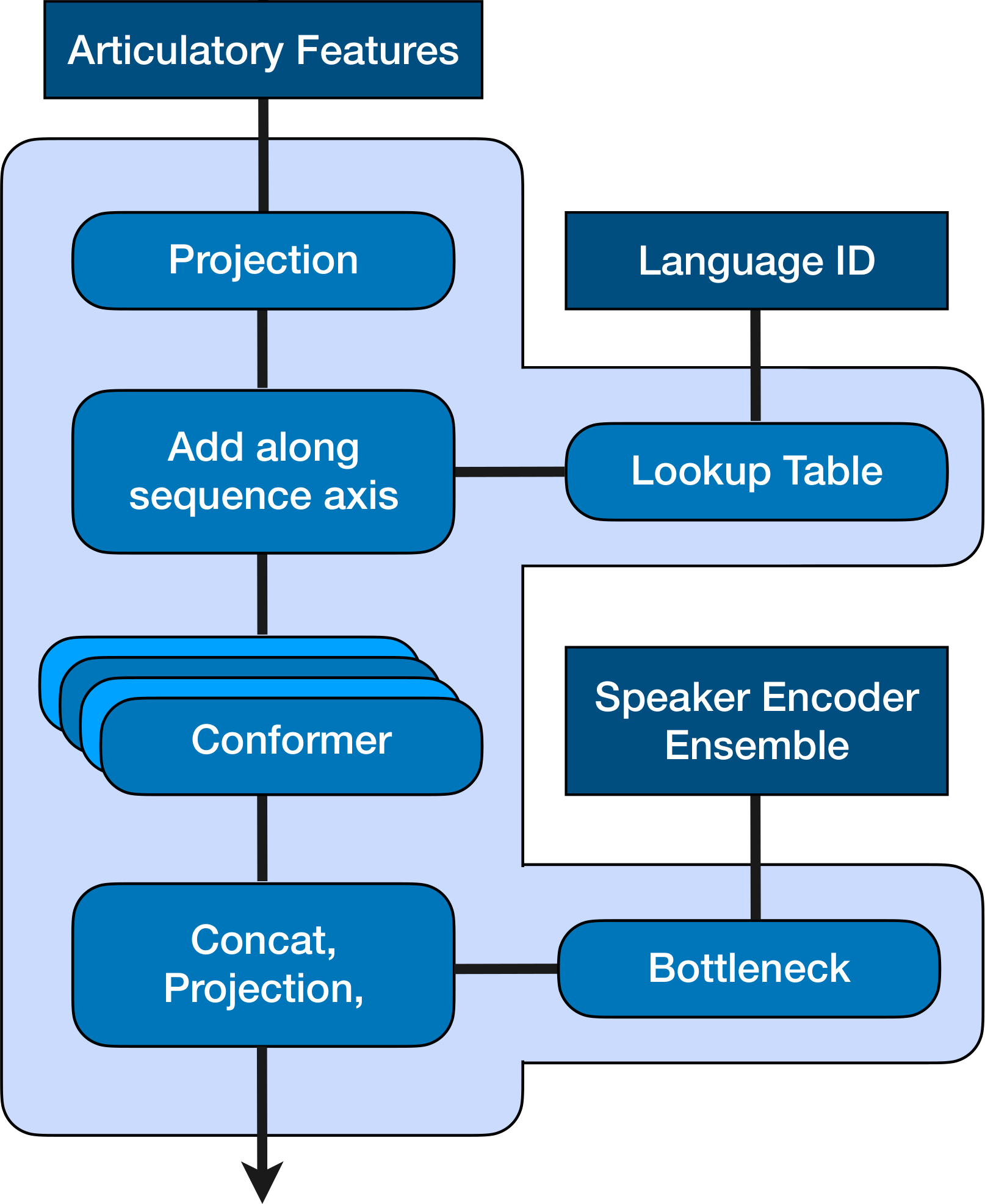}
	\caption{Overview of the encoder design. All of the projections project to the same dimensionality, which we chose to be 384. Round corners mean trainable. Conformer blocks include relative positional encoding.}
	\label{fig:enc_design}
\end{figure}

To handle the zero-shot multispeaker task, we condition the TTS on an ensemble of pretrained speaker embedding functions that consist of ECAPA-TDNN \cite{ecapa} and X-Vector \cite{xvect} trained on Voxceleb 1 and 2 \cite{Nagrani19, Nagrani17, Chung18b} using the SpeechBrain toolkit \cite{speechbrain} as suggested in \cite{anonymization}. Consistent with \cite{jia2018transfer} we find that the best ability to produce speech from voices unseen during training is achieved when injecting the speaker embeddings into the output of the encoder. First we bottleneck the speaker embeddings and apply the SoftSign function, as suggested in \cite{gibiansky2017deep}. Then we concatenate them to the encoder's hidden state and project them back to the size of the encoder's hidden state. At inference time, a speaker embedding of a reference audio can be used to make the synthesis speak in the voice of the reference speaker. An important trick we found is to add layer normalization right after the embedding is injected into the hidden state. This does not affect the synthesis of speakers seen during training, however it helps with unseen speakers.

In order to disentangle the languages from the speakers, we add an embedding for the language of the current sample along the sequence axis to the phoneme embedding sequence at the start of the encoder. This fits well to the intuition of a TTS encoder dealing with the text and the decoder dealing with the speech, since the text processing should not rely on speaker information, as a text does not have an inherent speaker. So we infuse the language information at the text stage and the speaker information at the speech stage of the model's information flow. Since, unlike the amount of possible voices, the amount of languages in the world is finite, we simply use an embedding lookup table to get embeddings of languages which receive their meaning purely through backpropagation during training. A text based language embedding could allow for zero-shot language adaptation, which we plan to investigate in the future. An overview of the multilingual multispeaker encoder is shown in Figure \ref{fig:enc_design}.

To transform the spectrograms that the FastSpeech 2 based synthesis produces into a waveform, we make use of the HiFi-GAN architecture \cite{kong2020hifi} as implemented in the IMS Toucan toolkit \cite{lux2021toucan}. As is shown in \cite{liu2021delightfultts}, neural vocoders can do super-resolution as well as spectrogram inversion. We apply the same trick to transform the 16kHz spectrograms the synthesis produces into 48kHz waveforms.

\subsection{Input Representation}
To make the use of multilingual data with only partially overlapping phonemesets easier, we represent the inputs to our system as articulatory feature vectors rather than identity based vectors, the same as is introduced in \cite{lux2022laml}. On top of this, we add an additional mechanism to deal with the multilinguality of the data.

Word boundaries are something that in most languages is very clearly visible in text. In spoken form however, word boundaries do not cover their own segment, but are instead only noticeable through cues in pitch and energy. This is why in TTS, word boundaries are usually removed. However we believe that in a multilingual setting, it is important to make the TTS model aware of word boundaries. We assume that this helps the model learn to distinguish how morpheme boundaries work in each language individually, as this is something that rarely holds across languages.

\begin{figure}[t]
	\centering
	\includegraphics[width=\linewidth]{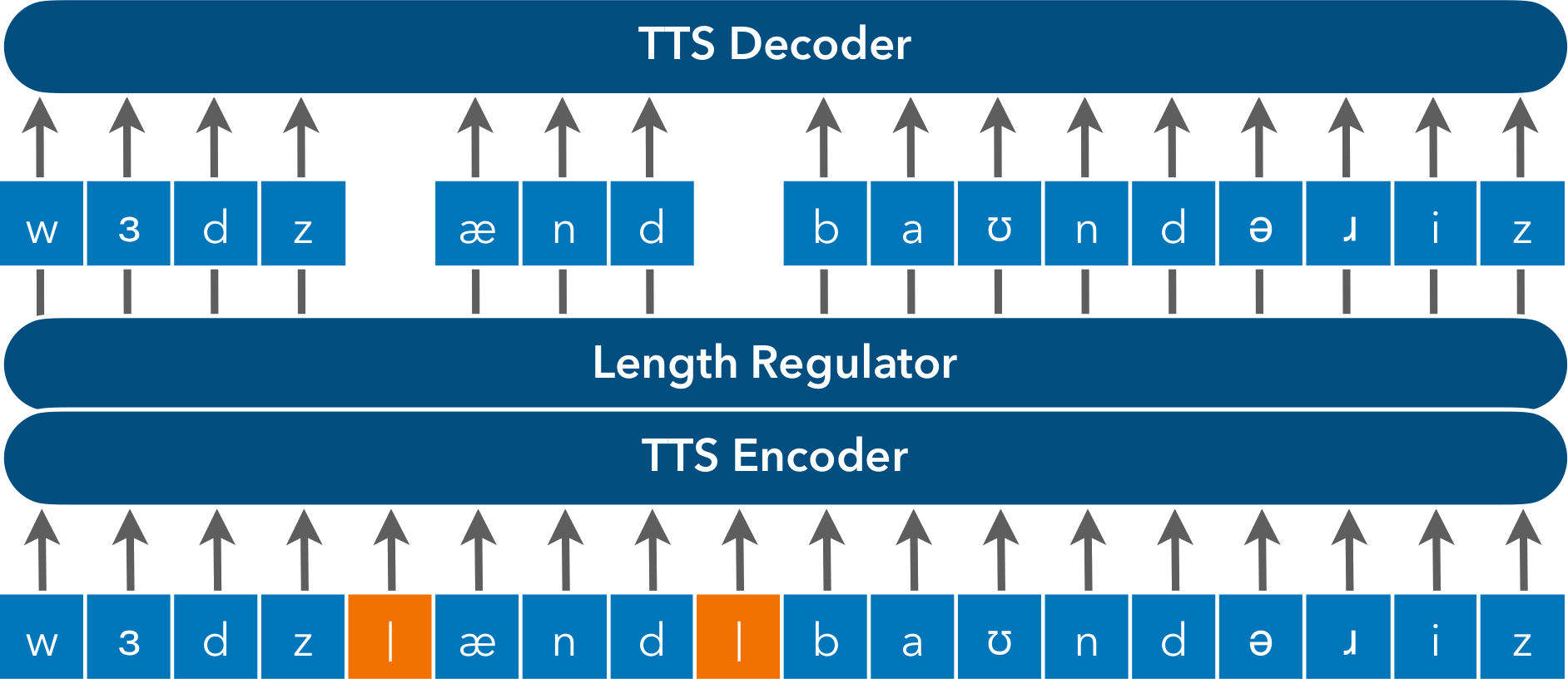}
	\caption{Example of the information flow of phonemes through the text encoder and speech decoder. The word boundaries (orange) are used in the encoder to contextualize the phoneme encoding, due to the length regulator however they do not reach the decoder.}
	\label{fig:bound}
\end{figure}

In our design, word boundaries are considered in the encoder of the TTS model, which intuitively corresponds to the encoding of the text, in which word boundaries do exist on the surface level, but not in the decoder, which intuitively corresponds to the decoding of the speech, where word boundaries are deeply embedded in the prosody as boundary tones. We achieve this by simply keeping track of the indexes of the word boundaries throughout the encoder and overwriting their predicted durations to be always zero. The upsampling mechanism in the length regulator will then remove their encoded vectors from the sequence as the information is passed to the decoder, while it was still available as contextual information in the encoder. This is illustrated in Figure \ref{fig:bound}. It is to be noted that as polar opposite to word boundaries, pauses do exist in speech, but not necessarily in text. For that reason, we treat pauses as separate units from word boundaries. Pauses receive a non-zero duration in the encoder and have their own spectrogram frames associated to them, unlike the word-boundaries. To detect pauses in the text, we use occurrences of commas and dashes in the text as a simple heuristic. This heuristic works in surprisingly many languages. Sentence marks like the question mark, the exclamation mark and the full stop are also treated as separate units, because they hold prosodic significance, even though they are mostly realized as a pause on the time axis.

\subsection{Data Preparation}
Furthermore we average the energy and pitch values extracted from the gold-audio over the spectrogram frames that belong to a single phoneme according to the alignment. This is introduced in FastPitch \cite{lancucki2021fastpitch} and allows for great controllability, but also makes model training more robust against low-quality data, which is an important feature for dealing with multilingual data since its quality greatly varies over the languages. 

Due to our reliance on spectrogram frames with their energy and pitch values being attributed to the correct phoneme, we make use of a lightweight self-contained aligner. We train this aligner as an automatic speech recognition system (ASR) using CTC \cite{graves2006connectionist}
and an L$_1$ reconstruction loss of its inputs and the outputs of an auxiliary TTS that backtranslates the frame-wise ASR predictions to a spectrogram inspired by \cite{perez2021vrain}. Alignment is then found by ordering the posteriograms of the ASR by the phonemes which we expect and then performing monotonic alignment search from start to end \cite{kim2020glow} using the efficient implementation from \cite{badlani2022one}. This aligner was introduced and is further described in \cite{cloning}.

\subsection{Training Procedure}
To train the TTS we make use of the LAML procedure \cite{lux2022laml}, which means that we treat different languages as tasks from a meta learning perspective. In order to solve all of these tasks simultaneously, an initialization point is iteratively refined to take fewer steps to get close to a good solution for each task. Such an initialization point that is well suited for all tasks seen in training is usually also suitable for unseen tasks (i.e. unseen languages in our context). 
To achieve this with TTS, we calculate the loss for one batch per language and sum them up. The samples from each language that go into each batch are chosen randomly, so the speakers are mixed throughout, resulting in also the ability to finetune to specific speakers on tiny amounts of data. 

Since phonemes should in theory be language agnostic, we also train the aligner on a massive amount of multilingual and multispeaker data described in section \ref{sec:data} following the same procedure resulting in low-resource finetuning capabilities. 

With regards to the vocoder we find that it can not only perform spectrogram inversion and super-resolution, but also slight speech enhancement. We inject random noise with a signal-to-noise ratio of 5db into the spectrogram for every tenth sample to increase the robustness of the vocoder against some noise in the synthesis induced by mixed quality data in some languages.

\section{Experiments}

\subsection{Data Used}
\label{sec:data}
In our experiments we use a variety of speech datasets with accompanying text labels in a total of 12 languages. The total amount of hours per language used is shown in parentheses in the following. For English (85h), we use the  Blizzard Challenge 2011 dataset \cite{king2011blizzard},  LJSpeech  \cite{ljspeech17}, LibriTTS \cite{zen2019libritts}, HiFi-TTS \cite{bakhturina2021hi} and VCTK \cite{veaux2017superseded}. For German (80h) we use the HUI-Audio-Corpus-German \cite{puchtler2021huiaudiocorpusgerman} and the Thorsten corpus \cite{muller_thorsten_2021_5525342}. Spanish (30h) includes the Blizzard Challenge 2021 dataset \cite{ling2021blizzard} and the CSS10 dataset \cite{css10}, from which we also use the Greek (4h), Finnish (11h), French (39h), Russian (21h), Hungarian (10h) and Dutch (34h) subsets. The Dutch and French subsets of the Multilingual LibriSpeech \cite{pratap2020mls} are also included, as well as its Polish (20h), Portuguese (25h) and Italian (30h) subsets. Greek, Finnish, Russian and Hungarian each only have a single speaker.
To have a high variety of data, but keep the computational cost manageable, we only use a maximum of 20,000 randomly chosen samples per corpus.

\subsection{Experimental Setup}
To verify our first contribution, we exclude German, Dutch and English data (Germanic languages) from the pretraining and then finetune a model on randomly chosen samples from a single speaker which add up to a total duration of just 5 minutes of German speech. We do the same with excluding Russian and Polish (Slavic languages) from the pretraining and then finetune on 5 minutes of Russian speech. In the evaluations we will refer to these models as the low-resource (LR) models. The two languages were chosen to simulate a low-resource scenario, rather than using an actual low-resource language, to still be able to get reliable and accurate measures on intelligibility and naturalness. We compare the two LR models to human speech as well as a single speaker model trained on 29 hours of German and 21 hours of Russian respectively. These models will be referred to as the high-resource (HR) models in the evaluation. Since the aligner and the vocoder are speaker and language agnostic, we exclude the Germanic and Slavic languages from their training and do not finetune them at all.

\paragraph{Intelligibility} To assess intelligibility, we calculate the phone-error-rate (PER) of the German and Russian IMS-Speech \cite{denisov2019ims} ASR systems on 3000 unseen sentences. This includes the case of an unseen speaker in the LR models. 

\paragraph{Naturalness} 
To verify the naturalness, we conduct a mean opinion score (MOS) study in which human raters give scores on a scale from 1-5 to 10 samples of human, LR and HR speech. For the case of German, we consider the HR model the upper bound, since the data is very high quality. Also, in this case the two largest and cleanest subsets of data were removed from the pretraining. So for German, we are investigating how close we can get to the performance of a very strong system. For Russian however, we can benefit from the high-quality pretraining that is met with less high-quality in-domain data and aim to even outperform the HR system.

\paragraph{Speaker Transfer} 
To verify our second contribution, we will measure the cosine similarity of speaker embeddings derived from synthetic speech to the embeddings derived from the human references used across all languages, including those which have seen only one speaker during training and the LR models from the previous experiment. A low standard deviation across all languages for each speaker (including the LR models) would indicate that the zero-shot multispeaker TTS properties are shared across all languages.

\paragraph{Word Boundaries}
The impact of the word boundaries can be mostly found in the intonation, but this includes cases where the intonation leads to incorrect phrasing and thus also incorrect word boundaries in the output. To verify their importance, we run the intelligibility experiment with a different configuration: We evaluate word-error-rate (WER) instead of PER and we only evaluate the German models, since the data quality is higher in that one, which gives us more reliable results. We compare each  model to a version that is trained completely analogous, but without word boundaries in the input. Since the HR models are monolingual, we hypothesize to see no change in WER, but an increase for the LR models, when the word boundaries are removed.

\paragraph{Accent Transfer}
To investigate the impact of the language embedding on its own, we focus on the languages which have only seen a single speaker during training, which are Greek, Russian and the two LR models. In these cases, it might be possible that the model has learned to associate the language with the voice of the speaker, since they always co-occur. We measure whether the cosine similarity to a target speaker in each of the other languages changes if we change the language embedding to one of the single-speaker languages. A small deviation would mean, that the language embedding does not affect the voice of the speaker, which is what we desire.  

\section{Results}

\subsection{Intelligibility}

The PERs of the different TTS systems are reported in Table \ref{wer_results}. The single speaker model for German almost matches the intelligibility of the human voice, indicating a very strong baseline. While the PER of the model trained on 5 minutes of a male German voice is worse relative to the single speaker model, the low absolute PER still indicates good intelligibility. When exchanging the speaker embedding for that of a female speaker, the PER increases further. This might be caused by the exclusion of the most varied and clean parts of the training data from the pretraining for this experiment, which reduces the overall quality for certain voices. It might however also simply be caused by the voice itself. Unfortunately, we do not have the same 3000 samples spoken by another speaker to investigate the impact of the voice on its own.

The Russian LR model also has a worse PER compared to human speech and the HR baseline. Looking into the cases where the LR model performed worse than the HR model, we mostly find near-misses, like producing the unvoiced variant of a consonant rather than the voiced variant. So while the small amount of data used paired with the lower quality of the finetuning data certainly negatively impacts the intelligibility, it is not as bad perceptively as the scores seem at first. Interestingly the impact of using a very different speaker embedding does not affect the PER significantly in this case. We assume this is because of the more diverse pretraining data that this model has seen.

\begin{table}[h]
	\centering
	\begin{tabular}{| l | l | l | c |} 
		\noalign{\hrule height 1pt}
		\textbf{Language}        & \textbf{Speech Type}                & \textbf{Voice} & \textbf{PER} \\
		\noalign{\hrule height 0.5pt}
		\multirow{4}{*}{German}  & Human  \hspace{.4cm}                & Male           & 3.58\%       \\   
		\cline{2-4}
		                         & TTS - HR                 & Male           & 3.59\%       \\ 
		\cline{2-4}
		
		                         & \multirow{2}{*}{TTS - LR} & Male           & 4.34\%       \\
		\cline{3-4}
		                         &                                     & Female         & 5.91\%       \\
		\noalign{\hrule height 0.5pt}
		\multirow{4}{*}{Russian} & Human                               & Male           & 7.65\%       \\
		\cline{2-4}
		                         & TTS - HR                 & Male           & 9.22\%       \\
		\cline{2-4}
		                         & \multirow{2}{*}{TTS - LR} & Male           & 12.32\%      \\
		\cline{3-4}
		                         &                                     & Female         & 12.64\%      \\
		\noalign{\hrule height 1pt}
	\end{tabular}
    \caption{\label{wer_results}
	PER of an ASR trained for the corresponding language. Reference speaker for LR speech is varied. The same 3000 samples are used to calculate each PER.}
\end{table}

\subsection{Naturalness}

\begin{table}[t]
	\centering
	\begin{tabular}{| l | l | c | c |} 
		\noalign{\hrule height 1pt}
		\textbf{Language}        & \textbf{Speech Type} & \textbf{MOS} & $\sigma$ \\
		\noalign{\hrule height 0.5pt}
		\multirow{3}{*}{German}  & Human  \hspace{.4cm} & 4.57 & $\pm0.69$        \\
		\cline{2-4}
		                         & TTS - LR   & 3.06 & $\pm1,35$         \\
		\cline{2-4}
		
		                         & TTS - HR  & 3.35  & $\pm1.02$      \\
		
		\noalign{\hrule height 0.5pt}
		\multirow{3}{*}{Russian} & Human                & 4.37 &  $\pm0.86$       \\
		\cline{2-4}
		                         & TTS - LR   & 3.57 & $\pm1.25$        \\
		\cline{2-4}
		                         & TTS - HR  & 2.07   &  $\pm1.02$    \\
		\noalign{\hrule height 1pt}
	\end{tabular}
    \caption{\label{mos_results}
	Mean opinion scores by human raters. All synthetic samples within a language are generated in the same voice.}
\end{table}

For the studies on the naturalness, we received a total of 330 ratings per speech type from 33 raters in German and 140 ratings per speech type from 14 raters in Russian. The results are shown in Table \ref{mos_results}. Considering that the setup for the German LR TTS is the most difficult, the model achieves a MOS that is surprisingly close to that of the baseline trained on 350 times more data, especially when considering the standard deviations, which indicate a large overlap in ratings. There is a rather large gap between the absolute values for human speech and synthetic speech, which is likely due to the very high quality of the human samples causing the raters to compare samples rather than rate them independent of each other. This causes even small imperfections to trigger a strong aversity. For Russian, the LR system even significantly outperformed the baseline trained on 250 times more data. We suspect that the mixed quality of samples in the Russian corpus (i.e. multiple different microphones and recording environments used) caused the single speaker model to not learn a consistent voice. It is however not a weak model, as the good performance on the intelligibility experiment confirms. In our interpretation, this shows that the pretraining can effectively leverage vast amounts of high-quality data in high-resource languages to perform well in underresourced languages.

\subsection{Speaker Transfer}
In preliminary experimentation we found that finetuning on the 5 minutes of data alone leads to rapid overfitting and the model loses its zero-shot multispeaker TTS capabilities. To prevent this, we finetune by including the small dataset into the LAML training procedure and train jointly for 5,000 batches. Further we found that when training with just one language per batch, the model does not converge to a usable state, whereas combining all languages to equal amounts in each batch (i.e. the LAML procedure) converges in just 60,000 steps, which shows the necessity of using LAML for this setup.

\begin{table}[h]
	\centering
    \resizebox{\linewidth}{!}{
	\begin{tabular}{| l | c  | c || l | c  | c |} 
		\noalign{\hrule height 1pt}
		&$\varnothing$&$\sigma$&&$\varnothing$&$\sigma$\\
		\noalign{\hrule height 0.5pt}
		English                       & 0.81                                    & 0.02   & 
	  Dutch                         & 0.79                                    & 0.03   \\ 
		German                        & 0.86                                    & 0.02   & 
		Finnish                       & 0.79                                    & 0.02   \\ 
		French                        & 0.85                                    & 0.01   & 
		Greek                         & 0.82                                    & 0.03   \\ 
		Hungarian                     & 0.77                                    & 0.04   & 
		Italian                       & 0.71                                    & 0.03   \\ 
		Portuguese                    & 0.75                                    & 0.03   & 
		Polish                        & 0.71                                    & 0.03   \\ 
		\textbf{Russian LR}           & 0.80                                    & 0.03   &
		Spanish                       & 0.81                                    & 0.03   \\ 
		\textbf{German LR}            & 0.81                                    & 0.03   &
		Russian                       & 0.79                                    & 0.03   \\ 
		\noalign{\hrule height 1pt}
	\end{tabular}
    }
 \caption{\label{tab:spk_sim}
	Cosine similarities of speaker embeddings of synthetic samples spoken in all 12 languages compared to the speaker embedding of the human reference speaker. Two utterances of the same human speaker leads to a similarity of 0.87 on average, defining an upper bound. $\varnothing$ is the average within-speaker similarity, $\sigma$ is standard deviation of the within-speaker similarity.}
	
\end{table}

\begin{figure}[t]
	\centering
	\includegraphics[width=\linewidth]{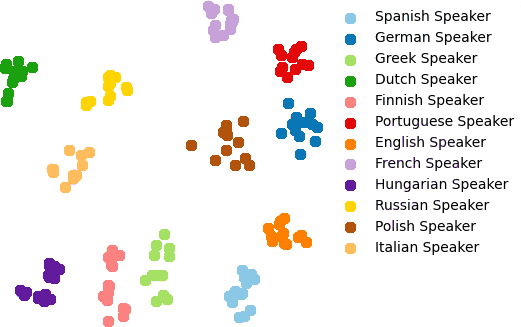}
	\caption{Visualization of speaker embeddings for 12 unseen speakers (1 speaker per language) each speaking 2 sentences in 12 different languages + the respective human speech reference. Each color corresponds to one speaker. Each point in a certain color is spoken in a different language.}
	\label{fig:spk_embs}
\end{figure}

Table \ref{tab:spk_sim} shows the average similarity that samples spoken in all 12 languages we investigated achieve compared to their human reference. The language column refers to the language of the speaker that the reference was taken from. A low standard deviation means, that the voice sounds similar regardless of the language it is currently speaking, indicating a good disentanglement of speakers and languages.
 While table \ref{tab:spk_sim} shows that the cloning of the speaker identity worked in some cases nearly perfect (German, French), there were also some cases where they didn't work as well (Italian, Polish). Investigating whether the language had an impact on this however showed, that the low scores are only due to the specific speakers which we randomly chose as the reference for those languages. Other speakers speaking either of those languages produced much higher similarities with their synthetic counterparts. So how well a voice can be cloned depends on the voice, but not on the language. The overall low standard deviations furthermore indicate that the speaker identity is consistent across all languages, regardless of which voice in which language is used as the reference.  For the LR variants included in this table, a different speaker than was seen in the training is used. The high similarity and low standard deviation indicates that the level of fulfillment of the zero-shot multispeaker TTS task exhibited by the full model is still present in the LR models. The results are supported by the visualization in Figure \ref{fig:spk_embs}. The clusters shown are linearly separable, indicating distinct speaker identities despite the switches in languages and high similarity to the human reference across all languages, even the ones where only a single speaker was seen during training.

\subsection{Word Boundaries}
As can be seen from Table \ref{wb_res}, the models that are aware of where word boundaries should go perform significantly better at placing the correct prosodic cues to indicate word boundaries in the output in the multilingual scenario. The impact of the boundaries on the monolingual model are insignificant.

\begin{table}[h]
	\centering
	\begin{tabular}{| l | c |} 
		\noalign{\hrule height 1pt}
		\textbf{Model}        & \textbf{WER}    \\
		\noalign{\hrule height 0.5pt}
		LR multilingual with boundaries &  13.71\%       \\ 
		LR multilingual without boundaries &  19.83\%       \\ 
		\hline
        HR monolingual with boundaries &  11.32\%       \\ 
		HR monolingual without boundaries &  11.91\%       \\ 
		\noalign{\hrule height 1pt}
	\end{tabular}
    \caption{\label{wb_res}
	Impact of monolingual and multilingual German models being aware of word boundaries as measured by an ASR system in terms of WER.}
\end{table}

\subsection{Accent Transfer}
Table \ref{simsim} shows whether the language embedding impacts the voice that is produced.  While the change of the language embedding did not significantly impact the similarity to the target speaker, we discovered that the information about the language encoded in the language embedding can actually be used to control the accent of the produced speech completely independent of language and speaker.

\begin{table}[h]
	\centering
	\begin{tabular}{| l | c || l | c |} 
		\noalign{\hrule height 1pt}
		\textbf{Embed.}        & \textbf{$\Delta$Sim} & \textbf{Embed.}        & \textbf{$\Delta$Sim}    \\
		\noalign{\hrule height 0.5pt}
		Greek &  0.001       &
		German LR &  0.002       \\ 
	  Russian &  0.008   &
		Russian LR &  0.004       \\ 
		\noalign{\hrule height 1pt}
	\end{tabular}
    \caption{\label{simsim}
	Average deviation in cosine similarity from target speaker in each language when the language embedding is switched to a language with only a single speaker.}
\end{table}

\section{Discussion}


\paragraph{Language Embedding Investigation} The accent transfer has interesting implications on how the distribution of realizations of a phoneme shifts with each language, independent of the context, which can be investigated by synthesizing individual phonemes with only the language embedding changed. We find language typical patterns, even in the languages that have only been trained on 5 minutes of data. So it seems that very little data is enough to capture a lot about how a language is usually spoken.

\paragraph{Implicit Morpheme Vocabularies} 
Although word boundaries are not explicitly denoted as segmental units in speech, they still have considerable influence on the phonetic realization. Consider for example the phenomenon of velar softening, i.e. a velar plosive is realized as alveolar fricative when followed by a long or short i ([\textsci] or [\textipa{ay}]) in some contexts, such as in \textit{electri[k] $\rightarrow$ {electri[s]ity}}. This does however not hold across word boundaries as in \textit{electri[\textipa{k}] igniter}.
Another example where word boundaries cause changes in the phone sequence is the phenomenon of final devoicing: voiced obstruents become voiceless if they occur in word-final position e.g. the German word \textit{Hunde (dogs)} is pronounced [h\textupsilon nd\textschwa] in its plural form but in singular \textit{Hund} becomes [h\textupsilon nt]. Such rules are however highly dependent on the language. Final devoicing is for example observed in German, Dutch and Polish, but not in English or French.

While many of these language specific lexical rules are already captured by the phonemizer, the situation is different in cases where word boundaries are not reflected by the phone sequence itself but only in the intonation, such as in ['acid] $\rightarrow$ [ac'id+ic]. While in the latter, there is still a morpheme boundary after \textit{acid}, this is not a word boundary. This highlights the importance of differentiating between actual word boundaries and word-internal morpheme boundaries in order to produce correct intonation which is crucial for generating intelligible speech.

Monolingual TTS models actually seem to learn an implicit vocabulary of morphemes as well as an intuition in which contexts morpheme boundaries can denote a word boundary in the language they are trained in. But in the case of multilinguality, this vocabulary of morphemes is difficult to construct, because every language has different morphemes. Thus, since multilingual models face a more difficult task to identify morphemes, they struggle even more distinguishing morpheme from word boundaries. Even with the language embedding, it seems like this is a property that the TTS can no longer implicitly capture, at least not given small amounts of data. 

We especially observe this in compound-nouns in our model trained on German in a low-resource setting. A model without explicit word boundaries adds boundary tones in the middle of the word causing an unnatural intonation that reduces the intelligibility of the word. If the model is trained with word boundaries, even though there are no word boundaries within the composite-noun, the pronunciation becomes much more fluent with the intonation being consistent throughout the word. 

\begin{figure}[t]
	\centering
	\includegraphics[width=.8\linewidth]{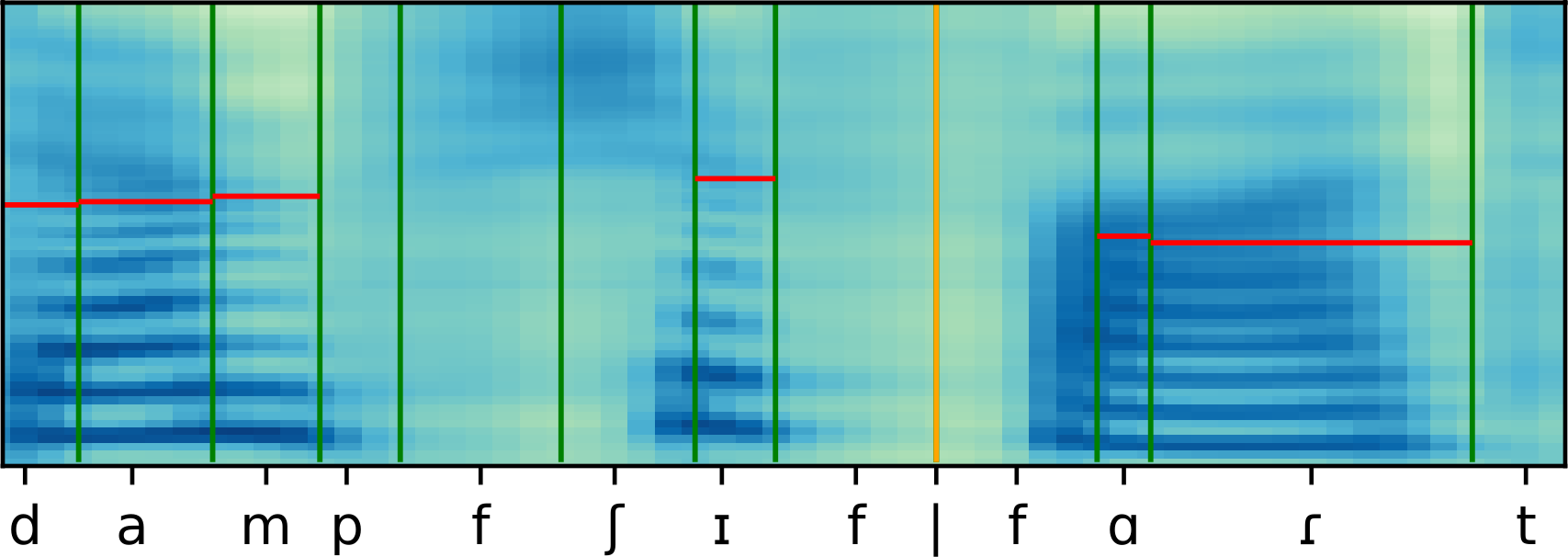}
	\includegraphics[width=.8\linewidth]{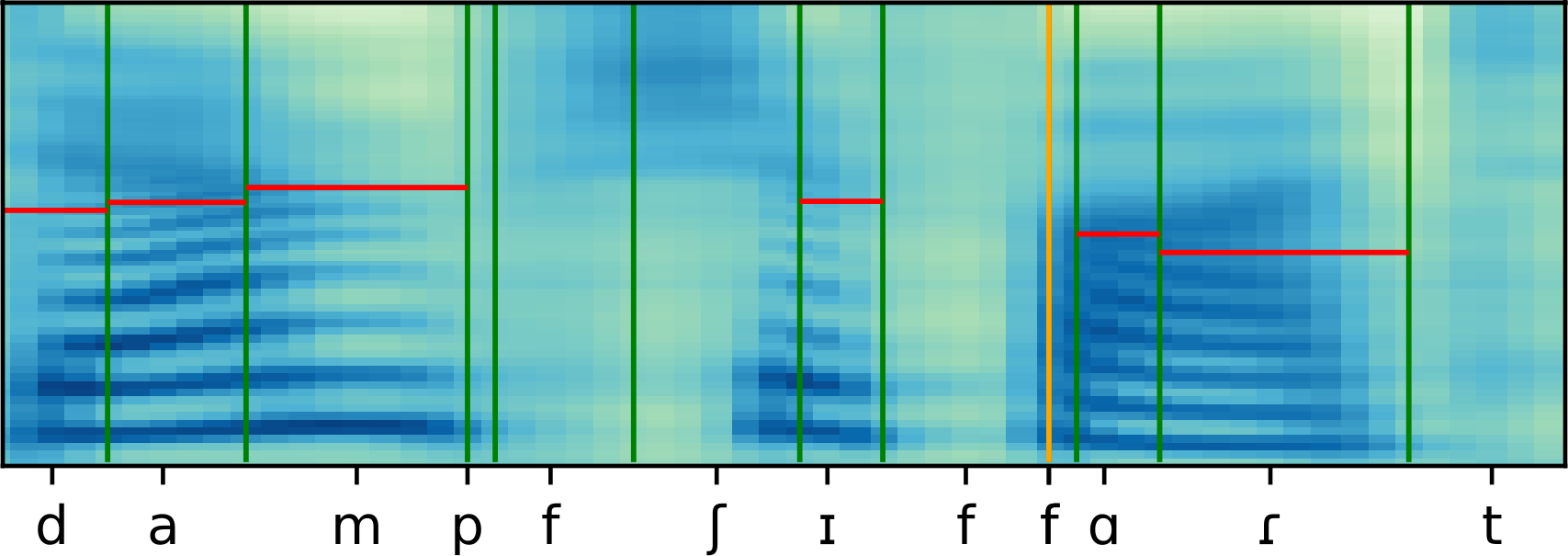}
	\caption{Spectrogram of the German noun-composite "Dampfschifffahrt" (steamboat ride) as produced by the word-boundary aware multilingual TTS (upper) and the multilingual TTS without word-boundaries (lower). Pitch predictions per phoneme are displayed in red, phoneme boundaries are displayed in green and the boundary between "steamboat" and "ride" in orange, which is however invisible to the models.}
	\label{fig:morpheme}
\end{figure}

Figure \ref{fig:morpheme} illustrates this with an example. It depicts spectrograms of a German word that consists of three parts: [\textipa{dampf}], [\textipa{\textesh \textsci f}] and [\textipa{f\textscripta\textfishhookr t}]. The components translate to English as steam, boat and ride. The proper phrasing within this word would be to combine the [\textipa{dampf\textesh \textsci f}] into one unit with the pitch being the highest on [\textipa{\textsci}] and a falling pitch towards the end of the word throughout [\textipa{f\textscripta\textfishhookr t}]. This is the case in the model that is aware of the word-boundaries. For illustration purposes, we include the boundary between steamboat and ride in the plot, the model however does not see this boundary as it happens in the middle of one word. The model which is unaware of the word-boundaries lowers its pitch already at the [\textipa{\textsci}] and lengthens the [\textipa{dampf}] part of the word. This makes the second instance sound as if the model was saying "steam boatride" rather than "steamboat ride".

We conclude that by simply making word boundaries explicit, the model no longer overestimates intonation phrase boundaries and boundary tones at every possible morpheme boundary.

\paragraph{Low-Resource Capabilities} Our experiments on low-resource scenarios show three major things: 1) it is possible to generalize into unseen branches in the phylogenetic tree of languages and reduce data-need even for languages with significant differences from the languages that have been trained on, which makes us hopeful that the direction of zero-shot learning to speak in a language is possible. 2) even from extremely little data in a target language, a lot of knowledge about the language can be abstracted. Language embeddings seem to encode language specific realizations of phones even when trained only on a few minutes of data. 3) the quality of data can be transferred across languages. Pretraining on high-quality data and then finetuning on low-quality data leads to a better model than when trained on much more of the low-quality data. This suggests that found data can be sufficient for TTS in a new language, because its quality can be improved by studio data in the pretraining.

\section{Limitations and Future Work}
While the LAML procedure is, as the name suggests, language agnostic, we only include European languages in our training and testing in order to get more reliable results with the resources for testing we have available. The state of the implementation with which the experiments were conducted cannot handle tonal languages, due to the non-segmental nature of tone. This limits the generality of our findings. Our open-sourced code has been updated in the meantime to be able to handle tone and lengthening properly. We plan to extend this work to include a much larger and much more diverse set of languages.

\section{Conclusion}
We show that through a simple encoder design coupled with a mechanism to encode word boundaries and the LAML training procedure, a low-resource capable multilingual zero-shot multispeaker TTS can be achieved. We are able to train a German and a Russian model on just 5 minutes of data each, which perform comparable or even better to single speaker models trained on 29 and 21 hours of data respectively. We further show that the ability to perform zero-shot multispeaker TTS is shared across languages, even those which have seen only 5 minutes of single speaker data. An additional side-effect is that the language embedding design in the encoder allows us to vary the accent of speech regardless of language of the input text and speaker.

\bibliography{main}
\bibliographystyle{acl_natbib}

\end{document}